%% file: main.tex
\documentclass{article}
\usepackage{spconf,amsmath,graphicx}
\ninept
\usepackage{amssymb}
\usepackage{balance}
\usepackage{caption}
\usepackage{subcaption}
\usepackage{amsmath}
\usepackage{amsfonts}
\usepackage{graphicx}
\usepackage{cite}
\usepackage[colorlinks=true, allcolors=blue]{hyperref}
\usepackage[dvipsnames]{xcolor} % package for colored text
\usepackage{ntheorem}

\input{macros}
\title{Mitigating Data Injection Attacks on Federated Learning}

\address{}
\name{Or Shalom$^1$\thanks{This work is partially supported by ISF grant 2197/22.}, Amir Leshem$^2$, Waheed U. Bajwa$^3$}

\address{$^1$shalomo7@biu.ac.il; $^2$amir.leshem@biu.ac.il; $^3$waheed.bajwa@rutgers.edu
\\
$^{1,2}$Faculty of Engineering, Bar-Ilan University, Ramat Gan 5290002, Israel
\\
$^3$Dept.\ of Electrical \& Computer Engineering, Rutgers University--New Brunswick, NJ 08854 USA
}

\begin{document}
\maketitle
\begin{abstract}
Federated learning is a technique that allows multiple entities to collaboratively train models using their data without compromising data privacy. However, despite its advantages, federated learning can be susceptible to false data injection attacks. In these scenarios, a malicious entity with control over specific agents in the network can manipulate the learning process, leading to a suboptimal model. Consequently, addressing these data injection attacks presents a significant research challenge in federated learning systems. In this paper, we propose a novel %technique 
approach to detect and mitigate data injection attacks on federated learning systems. Our mitigation %method
strategy is a local scheme, performed during a single instance of training by the coordinating node, allowing %the 
for mitigation during the convergence of the algorithm. Whenever an agent is suspected %to be 
of being an attacker, its data will be ignored for a certain period; this decision will often be re-evaluated. We prove that with probability one, after a finite time, all attackers will be ignored while the probability of ignoring a trustful agent becomes zero, provided that there is a majority of truthful agents. Simulations show that when the coordinating node detects and isolates all the attackers, the model recovers and converges to the truthful model. 
\end{abstract}
\section{Introduction}
\label{sec:intro}

Big-data processing capabilities have increased significantly over the past years due to the increasing volume and variety of data that is being generated and collected. Today, data has become an asset, and processing that data is required in many, if not all, the industries affecting our lives, such as healthcare, finance, transportation, manufacturing, and many more. With the increased need for data, a new need for the privacy and security of the data has risen \cite{price2019privacy,jain2016big}.

\begin{figure}
    \centering
    \includegraphics[width=0.45\textwidth]{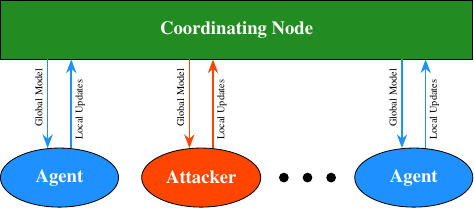}
    \caption{A simple federated learning system. Each agent %is performing 
    performs training on its private dataset; the local updates are then transmitted to a coordinating node, which returns a global model. Some of the agents may be malicious, meaning they might send unreliable updates to the coordinating node.}
    \label{fig:fl_simple}
%\vspace{-\baselineskip}
\end{figure}

Federated learning is a popular approach for collaboratively training machine learning models while preserving data privacy \cite{zhang2021survey,li2020federated}. %In this approach, rather than traditionally training a centralized model with a combined dataset, there are multiple independent agents, each with its private dataset, assumed to be independent of the other datasets, performing training on a local model. 
In this approach, instead of training a centralized model using a combined dataset as traditionally done, multiple independent agents train local models using their private datasets, which are assumed to be statistically independent. The agents %are exchanging 
exchange the local model parameters (i.e., weights and biases) with a centralized node coordinating the learning process, which in turn returns a new model (or model updates) to all the agents (cf.~Figure~\ref{fig:fl_simple}). Although the agents %are benefiting 
benefit from the training performed on other agents' data, the datasets are not shared and remain private.

Even though the federated learning model preserves privacy, it is vulnerable to various security threats, including data injection and poisoning attacks \cite{li2022detection,zhao2021federated,tolpegin2020data,fang2020local}, backdoor access \cite{bagdasaryan2020backdoor,yar2023backdoor}, gradient attacks \cite{geiping2020inverting} and many more \cite{lyu2020threats,mammen2021federated}. In data injection attacks, malicious participants (agents) inject false data into the training process to manipulate the global model. Detecting data injection attacks in %decentralized 
federated learning is %a challenging task, as the data is distributed across multiple devices and there is no central authority with access to the entire data set to monitor the training process. 
challenging due to the distributed nature of the data across multiple devices. Although a coordinating node oversees the training process, it does not have access to the complete dataset, limiting its ability to comprehensively monitor for these attacks. This has led to the development of various techniques for detecting and preventing data injection attacks, including model-based methods \cite{wu2018data,ravi2019detection}, anomaly detection techniques \cite{shalom2022localization}, Byzantine resilient methods \cite{xu2022byzantine,so2020byzantine,FangYangEtAl.ITSIPN22,yang2019byrdie}, and federated outlier detection \cite{ghosh2019robust,rodriguez2022backdoor}.  

%Previous works on the detection of data injection attacks in federated learning include, e.g., Tolpegin et al. \cite{tolpegin2020data} which proposed a PCA-based detector. 
Previous works on detecting data injection attacks in federated learning have made notable contributions. For instance, Tolpegin et al.\ \cite{tolpegin2020data} proposed a PCA-based detector. Based on the difference in agents' parameter updates, they %can
were able to separate the parameter updates into two clusters, one %of 
comprising malicious agents and the other consisting of trustworthy agents. Yar et al.\ \cite{yar2023backdoor} suggested a modified dual ``gradient clipping defense". Standard clipping defense is a %defense 
scheme specifically designed for data poisoning attacks, in which agents that send updates with norms too high are clipped. Yar et al.\ showed that in a dynamic network, a dual threshold clipping defense with one smaller threshold for neighboring agents and another larger threshold for the global model achieves better results.

In this paper, we present a novel detection method for data injection attacks in federated learning. The method is performed along the model training process and is based on evaluating the gradient of the updates of the participating agents, comparing this gradient to the coordinatewise median over the agents, and ignoring updates from suspicious agents. Considering the history of detections using a majority voting among the coordinating node's decisions, we can overcome false alarms and missed detections. We prove convergence to a truthful model with probability one, provided that data is independent and identically distributed (i.i.d.) among agents. %One might state that assuming the data is i.i.d might be limiting. 
It could be argued that the assumption of i.i.d.\ data might impose limitations. However, many theoretical papers use i.i.d.\ data for the theoretical analysis (see \cite{tolpegin2020data,achituve2022communication}. This has also been the case with signal processing adaptive algorithms where i.i.d.\ data %was 
is widely assumed. The approach taken generalizes the decentralized optimization for M-estimators in \cite{shalom2022localization} to the federated learning context. We also demonstrate by simulations that the proposed technique can overcome constant output and label-flipping attacks \cite{li2022detection,zhao2021federated,tolpegin2020data,fang2020local,bagdasaryan2020backdoor,yar2023backdoor,geiping2020inverting,lyu2020threats,mammen2021federated}, even when these attacks are hidden with partially truthful responses.

\section{Problem Formulation}
\subsection{Federated Learning}
The federated learning problem involves learning a model using agents' private data. %Each agent uses its private data to refine its model parameters; the local updates are transmitted to the coordination node, which then returns a global model. Due to the nature of this algorithm, only the model's parameters are being shared among the users. 
In this setting, each agent, using its private data, refines its model parameters. These local updates are then transmitted to the coordinating node, denoted as agent $0$, which synthesizes them into a global model. This process ensures that only the model's parameters, not the data itself, are shared among agents.

Consider a dataset $\mathcal{D}$ labeled with labels from a set $C$. %Each agent refines its model parameters iteratively during the learning phase using the labeled dataset. Typically, the goal is to minimize an objective function using a gradient descent approach:
Within this framework, agents $1, \ldots, N$---referred to as the edge agents---iteratively refine their model parameters during the learning phase using this labeled dataset. The objective typically involves minimizing a function using a gradient descent approach:
\begin{equation}
\min_W F(W), \text{ where } F(W) := \sum_{k=1}^N p_k F_k(W),
\end{equation}
%given that $N$ is the number of participating agents
where $N$ represents the number of participating agents, $\sum_k p_k=1$, and $F_k$ is the local empirical risk function for the $k$-th agent. %Typically, $p_k=\frac{1}{N}$ but different values can be used to prioritize the risk of certain agents.
Although $p_k=\frac{1}{N}$ is common, varying these values can prioritize the risk of certain agents. %Let agent $0$ be the coordinating node and agents $1,\ldots,N$ are the edges agents. In each time step, each edge agent performs learning on its private dataset. Once all the agents are done with the learning phase, they broadcast their model parameters to the coordinating agent. Each agent receives the average model parameters computed by the coordinating agent. 
At each time step, after completing their learning phase, the edge agents broadcast their model parameters to the coordinating node, which then computes and returns the averaged model parameters to all agents.

\subsection{Data Injection Attacks}
\label{sec:attacks}

The federated learning approach was previously shown to achieve excellent results \cite{haddadpour2019convergence,nguyen2020fast}, particularly when all collaborating agents share the same goal. However, consider a scenario where some agents participating in federated learning are malicious. We use the following notation: Denote the set of attackers $A \subset {1,\ldots, N}$ and let $n_a = |A|$. We assume that $0 \le n_a<N/2$, and $n_t=N-n_a$ is the number of trustworthy agents.

%Since the attacking agents must be coordinated to achieve a desired false model, we assume w.l.o.g that we have a single malicious agent, say agent $a$, participating in the joint model training. 
In this context, since attacking agents must coordinate to achieve a false model, we can assume, without loss of generality, a single malicious agent, say agent $a$, influencing the joint model training. Agent $a$'s goal is to manipulate the joint model training, thus preventing it from performing successfully by transmitting false parameters, voiding the convergence of the model around an optimal point, and steering it towards a false model with predetermined performance. This agent is capable of executing various destructive attack schemes, such as ``label flipping attack", ``constant output attack", ``randomized attack" and %many 
more (see \cite{li2022detection,zhao2021federated,tolpegin2020data,fang2020local}). 

In a randomized attack, the attacker broadcasts a random set of model parameters back to the other agents%. This attack is powerful since it quickly destroys the learned model performance.
, a tactic that rapidly degrades the learned model's performance. %However, such an attack is relatively easy to detect since the malicious agents' responses significantly differ from the trustworthy ones. 
However, this type of attack is relatively easy to detect, as the malicious agents' responses stand out significantly from those of the trustworthy ones. The constant output attack aims to steer the system towards a model that %always
consistently classifies a single class $c$ irrespective of the data. %The last type of attack which is harder to detect performs a label flipping by choosing a specific permutation of the labels, and responding as if this permutation is applied to the output of the training data or by sending a model trained to identify the permuted class values, e.g., an attack on MNIST can be using a model classifying class $c$ as class  $(c+2)~\text{mod}~10$ rather than the true value $c$ for any given training data point.
Lastly, the more challenging to detect label flipping attack involves choosing a specific permutation of the labels. The attacker responds as if this permutation has been applied to the training data's output or by sending a model trained to recognize the permuted class values. For example, in an attack on MNIST, this could involve using a model that classifies class $c$ as class $(c+2)~\text{mod}~10$ rather than its true value $c$ for any given training data point.

A stronger type of attack would try to hide the previous attacks by combining a false model and a true one. This reduces the statistical discrepancy between the malicious agents' response and the other agents' response, by adding a bias to the reported model. While the attacker's main goal is to manipulate the joint model parameters and prevent convergence to a steady optimal point, it also has a secondary goal of remaining hidden and disguising the attack. In this strategy, let $W_{a,r}(t)$ %be the local model parameters obtained
represent the model parameters that the attacker, posing as a regular agent, would have updated by reliably updating the model $W(t-1)$ provided by the coordinating agent with correct data at time $t$, while $W_{a,f}$ is a pre-trained false model, classifying labels according to the attacker's desired attack scheme.\footnote{For simplicity of notation we assume $W_{a,f}$ is time independent, but it can also be dependent.}

%Formally, this can be described as follows: 
Building on this concept, the attack can be formally described as follows: A malicious agent $a$ responds at time $t$ by sending %the model
\begin{equation}
	\label{eq:New Attack Model}
	\begin{split}
		W_a(t) := g(t)W_{a,r}(t) + (1-g(t))W_{a,f},
	\end{split}
\end{equation}
where $W_a(t)$ %refers to 
denotes the set of parameters transmitted by agent $a$ at time $t$, and $g(t)$ is a time-varying mixing weight satisfying the following conditions:
\begin{itemize}
	\item For all $t\geq 0$, $0\leq g(t) \leq 1$,
 	\item $g(0) \equiv 1$, $\lim_{t\to\infty}g(t) = 0$, and
	\item $g(t)$ decreases over time, i.e., $g(t+1) \le g(t)$.
\end{itemize}
For instance, delaying the start %ing time 
of the attack to time $T_a$ %is equivalent to 
can be achieved by setting $g(t)=1, 0\le t\le T_a$. %The monotonicity assumption might be relaxed but is required for the convergence to the attacker's desired model.
While the monotonicity of $g(t)$ can be relaxed, it is essential for ensuring convergence to the attacker's desired model.

Note that this attack scheme is realistic as the attacker has no access to other agents' datasets and it can't manipulate the model learning process or the parameters aggregation done by the coordinating agent. The attacker is only capable of pre-learning a manipulated model and it doesn't rely on a specific neural network configuration, optimization function, or loss function.

\section{Attacker Detection and Avoidance}
\label{sec:localization}
In %this 
our proposed method, the coordinating agent compares the updates received from edge agents over time. The private datasets are assumed to be identically distributed and therefore if an agent is malicious and its model parameters update differently, it will stand out and be considered malicious.

To localize the attacker, we propose a low-complexity metric, computed over time by the coordinating agent once every $\Delta T$ updates. When the coordinating agent suspects an edge agent to be an attacker, it ignores its parameter updates for the next $\Delta T$ updates. %Denote 
Let $I_k=[(k-1)\Delta T+1, k\Delta T]$. %Denote 
Define the two hypotheses tested over the interval $I_k$:
\begin{equation*}
	\begin{split}
		&\mathcal{H}^0_{j,k} \text{ -- agent $j$ is trustworthy}. \\
		&\mathcal{H}^1_{j,k} \text{ --  agent $j$ is malicious}.
	\end{split}
\end{equation*}

The proposed detection metric for a given interval $I_k$, computed over time for agent $j$'s model parameters, is given by 
\begin{equation}
	\begin{split}
		\Delta U_{j} := \frac{1}{\Delta T}\sum_{t\in I_k} U_{j}(t) \mathop {\lessgtr}_{\mathcal{H}^1_{j}}^{\mathcal{H}^0_{j}} \delta_u\sqrt{N},
	\end{split}
	\label{localization_u}
\end{equation}
where
\begin{equation}
    U_{j}(t) :=  \| \Delta W_j(t) - \text{median}\{\Delta W_\ell(t) : \ell\in \{1,\ldots,N\} \setminus \{j\}\} \|_\infty.
\label{Equation_14_b}
\end{equation}
Here, the median is a coordinatewise operation, $\Delta W_j(t):=W_j(t)-W_j(t-1)$, and $\delta_u$ is a predefined threshold. The following lemma characterizes the probability of accurately identifying malicious agents using the metric $\Delta U_j$ under certain assumptions.
\begin{lemma}
Assume that the majority of agents are trustworthy. Furthermore, assume that data is sub-Gaussian and i.i.d.\ between agents and classes. There are values $\delta_u$ and $\Delta T$ for which $P_{FA}(I_k)<\frac{1}{2}<P_D(I_k)$ when detection is based on consecutive $\Delta T$ model updates, where $P_{FA}$ denotes the probability of a false alarm and $P_D$ denotes the probability of detection.
\end{lemma}
%The proposed detector allows for continuous operation, regardless of the convergence time of the joint model, using the insight above. 
The proposed detector facilitates continuous operation, unaffected by the convergence time of the joint model, drawing on the insights from the lemma. Let $d_k$ %be 
denote the outcome of %implementing 
applying (\ref{localization_u}) on interval $I_k$. 
This results in a sequence of decisions $d_1,d_2,\ldots$, where $d_i=1$ if the examined agent crosses the threshold, and $d_i=0$ otherwise. %By the end of each segment, $K \Delta T$  the coordinating agent ignores the input of all edge agents for which
At the conclusion of $K \Delta T$ updates, the coordinating agent assesses each edge agent's behavior. If an edge agent's average decision score over these $K$ intervals, calculated as
\begin{equation}
    \frac{1}{K}\sum_{k=1}^K d_k,%>\frac{1}{2},
        \label{eq:remove}
\end{equation}
exceeds \(1/2\), the agent's input is excluded for the next segment \(I_{K+1}\). Nonetheless, the coordinating agent continues to compute the statistics \eqref{localization_u} during this period. Furthermore, if 
\begin{equation}
    \frac{1}{K}\sum_{k=1}^K d_k<\frac{1}{2},
    \label{eq:insert_back}
\end{equation}
the agent is added back to the list of trustworthy agents. %The following lemma holds:
The validity of this approach is encapsulated in the following lemma:
\begin{lemma}
\label{new_lemma_6}
Assume we set $\delta_u$, $\Delta T$ such that for each $k$, $P_{FA}(I_k) < 1/2 < P_D(I_k)$. Then, with probability 1, there exists a sufficiently large $k_0$ such that the presented scheme (cf.~\eqref{localization_u}--\eqref{eq:insert_back}) ignores all the malicious agents after time $k_0\Delta T$, while ensuring that updates from all trustworthy agents are incorporated beyond this time. %and all trustworthy agent updates are used beyond this time.
\end{lemma}

The proof of Lemma \ref{new_lemma_6} is based on a sequence of decisions where for each interval $I_k$ of length $\Delta T$ the detector is applied to obtain a decision $d_k$. Then a majority among all prior decisions is used to decide whether to disconnect the agent. By the assumption $P_{FA}<\frac{1}{2}<P_D$ and the Borel Cantelli lemma the proof follows.

\section{Numerical Simulations}
In this section, we %present simulated attacks on decentralized learning and evaluate the performance of our detection algorithm. 
evaluate the effectiveness of our detection algorithm through simulated attacks on federated learning systems, utilizing the MNIST dataset for this purpose. %We use the MNIST data set. This data set is composed of images of handwritten digits $0,\ldots,9$ and the goal is to classify the correct digit.
The MNIST dataset comprises images of handwritten digits $0,\ldots,9$, with the objective being to accurately classify each digit. %Since the metric of interest is the classification error of the learned model, we use the average error of the learned model to evaluate the performance of the algorithm, rather than providing detection probabilities for the agents.
To gauge performance, we focus on the classification error of the learned model as our primary metric of interest. Consequently, we evaluate the algorithm's efficacy based on the average error rate of the learned model, rather than quantifying detection probabilities for individual agents.

\begin{figure}[t!]
     \centering
         \begin{subfigure}[b]{0.5\textwidth}
         \centering    
	\includegraphics[width=1\textwidth]{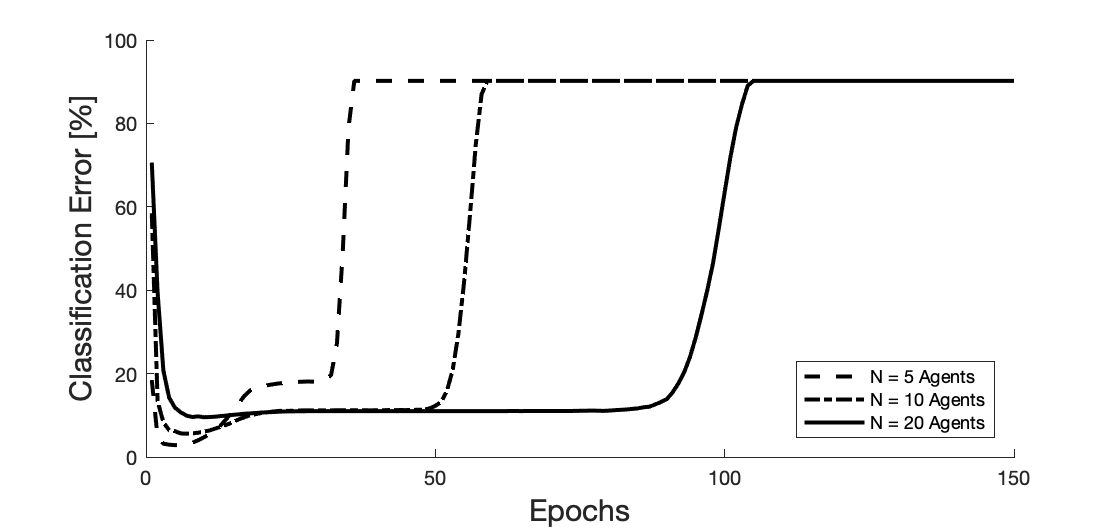}
         %\caption{Constant-Output attack by a single attacker.}
         \caption{Impact of constant-output attack without mitigation on classification error.}
     \end{subfigure}
     \hfill
     \begin{subfigure}[b]{0.5\textwidth}
         \centering
	\includegraphics[width=1\textwidth]{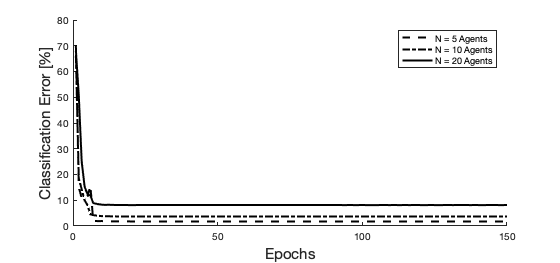}
         %\caption{Detection of a single attacker performing Constant-Output attack.}
         \caption{Efficacy of the proposed detection scheme against constant-output attack.}
     \end{subfigure}
        %\caption{A comparison of a Constant-Output attack by a single attacker on various network sizes with and without detection. In the plots, we see the classification error on the test dataset. The top plot shows an attacked network with no detection and the bottom plot shows an attacked network with detection.}
        \caption{Assessing Constant-Output Attack and Detection in Federated Learning: (a) Classification error under a single attacker across different truthful agent ratios, highlighting the attack's success without detection. (b) Effectiveness of the detection scheme, emphasizing network resilience and model accuracy against the attack.}
        \label{fig:ConstantOutputAttackDiffSizes}
\vspace{-\baselineskip}
\end{figure}

We performed 100 random cross-validation experiments, %where at each instance
with each involving 60,000 images to train the agents and 10,000 images for testing their performance. Each agent received a random set of $60000/N$ different images, %disjoint
ensuring they were distinct from the images of the other agents, for training. The number of participating agents included $N=5,10,20$, with one designated as an attacker. The attacker's mixing weight was \(g(t) = 1/\sqrt{t+1}\) and the integration time was \(\Delta T=5\). We implemented a 7-layer convolutional neural network (CNN):
\begin{center}
\begin{tabular}{ |c| c | c| c| }
\hline
 \textbf{Layer} & \textbf{Size} & \textbf{Input} & \textbf{Output} \\ 
 \hline
 Conv+Relu & 1x32x3x1 & 1x28x28 & 32x26x26 \\
 \hline
 Conv+Relu & 32x64x3x1 & 32x26x26 & 64x24x24 \\
 \hline
 MaxPool2D & 2x2 & 64x24x24 & 64x12x12 \\
 \hline
 Dropout + Flatten & $p=0.25$ & 64x12x12 & 1x9216 \\
 \hline
 FC + Relu & 9216x128 & 1x9216 & 1x128 \\
 \hline
 Dropout & $p=0.5$ & 1x128 & 1x128 \\
 \hline
 FC & 128x10 & 1x128 & 1x10\\
 \hline
\end{tabular}
\end{center}
where a `Conv' layer size is $({CH}_{in},{CH}_{out},kernel,stride)$.

\begin{figure}[t!]
     \centering
         \begin{subfigure}[b]{0.5\textwidth}
         \centering    
	\includegraphics[width=1\textwidth]{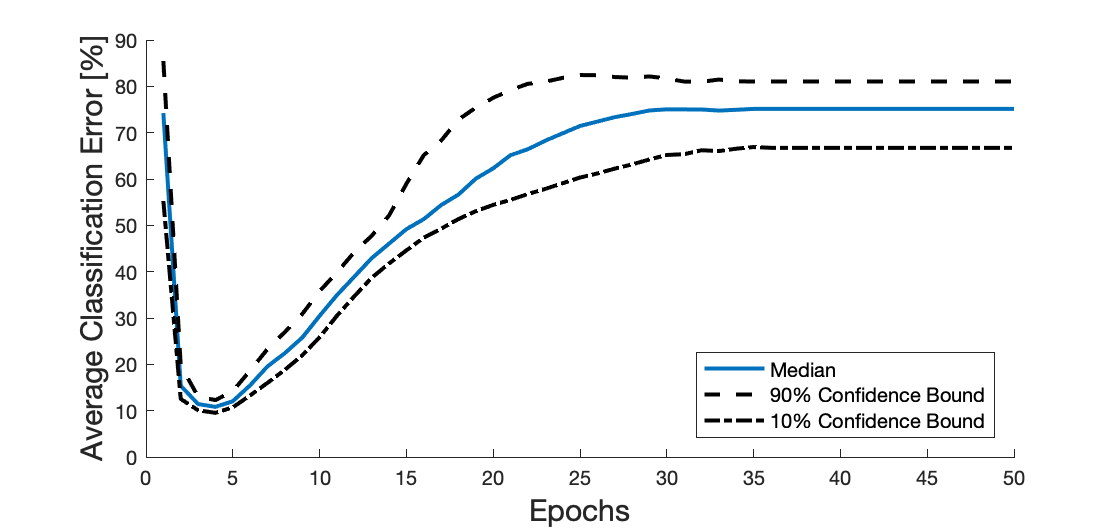}
         %\caption{$100$ Experiments of an attacked network.}
         \caption{Attack Impact Without Detection: $100$ Experiments.}
     \end{subfigure}
     \hfill
     \begin{subfigure}[b]{0.5\textwidth}
         \centering
	\includegraphics[width=1\textwidth]{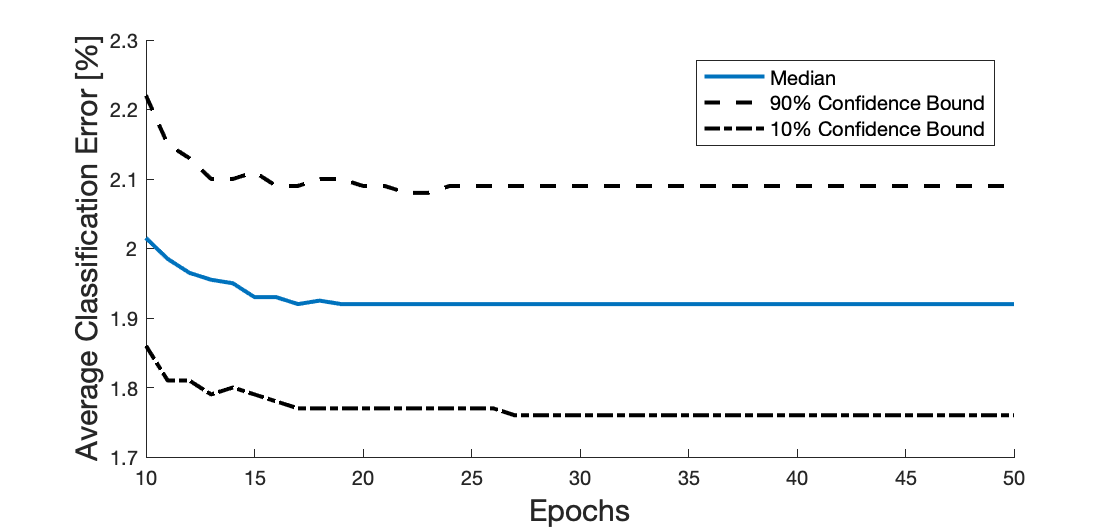}
         %\caption{$100$ Experiments of an attacked network with detection.}
         \caption{Network Resilience With Detection: $100$ Experiments.}
     \end{subfigure}
        %\caption{$100$ Experiments of Constant-Output attack by a single attacker with and without detection each on a network with $N=5$ agents. The above plots show the $10\%$ and $90\%$ confidence bounds for the model's average classification error.}
        \caption{Comparative Analysis of Constant-Output Attack: Showcasing $100$ experiments on a $N=5$ agent network, this figure highlights the model's classification error with and without the detection scheme [(a) and (b)]. Included are $10\%$ and $90\%$ confidence bounds, underscoring the attack's effect and the detection's efficacy.}
        \label{fig:ConfidenceBoundsExperiments}
%\vspace{-\baselineskip}
\end{figure}

\begin{figure}[ht!]
     \centering
         \begin{subfigure}[b]{0.5\textwidth}
         \centering    
	\includegraphics[width=1\textwidth]{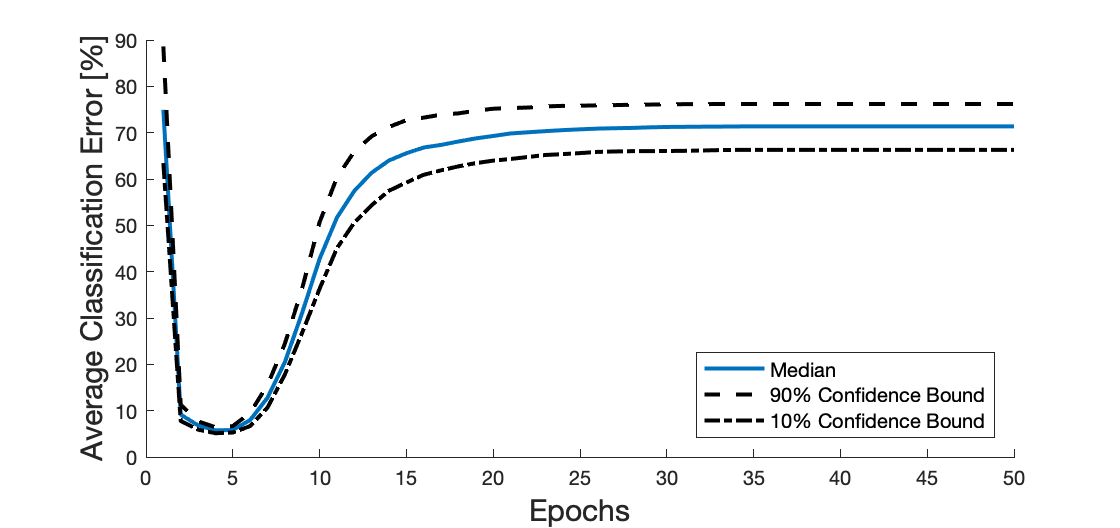}
         %\caption{$100$ Experiments of an attacked network.}
         \caption{Classification Error Without Detection (100 Experiments).}
     \end{subfigure}
          \hfill
     \begin{subfigure}[b]{0.5\textwidth}
         \centering
	\includegraphics[width=1\textwidth]{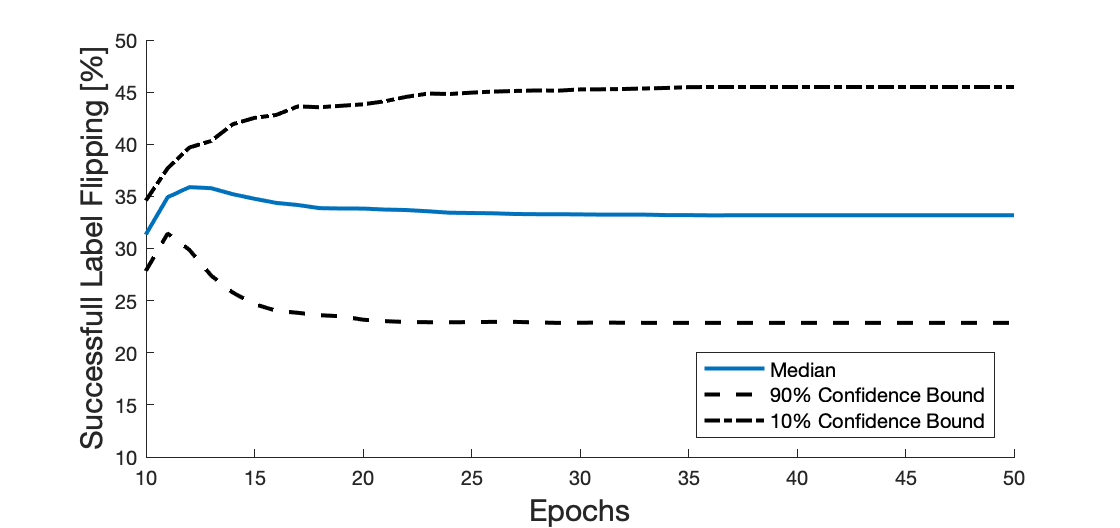}
         %\caption{$100$ Experiments of an attacked network, showing the average number of successful label-flips.}
         \caption{Label-Flip Success Rate Without Detection (100 Experiments).}
     \end{subfigure}
     \hfill
     \begin{subfigure}[b]{0.5\textwidth}
         \centering
	\includegraphics[width=1\textwidth]{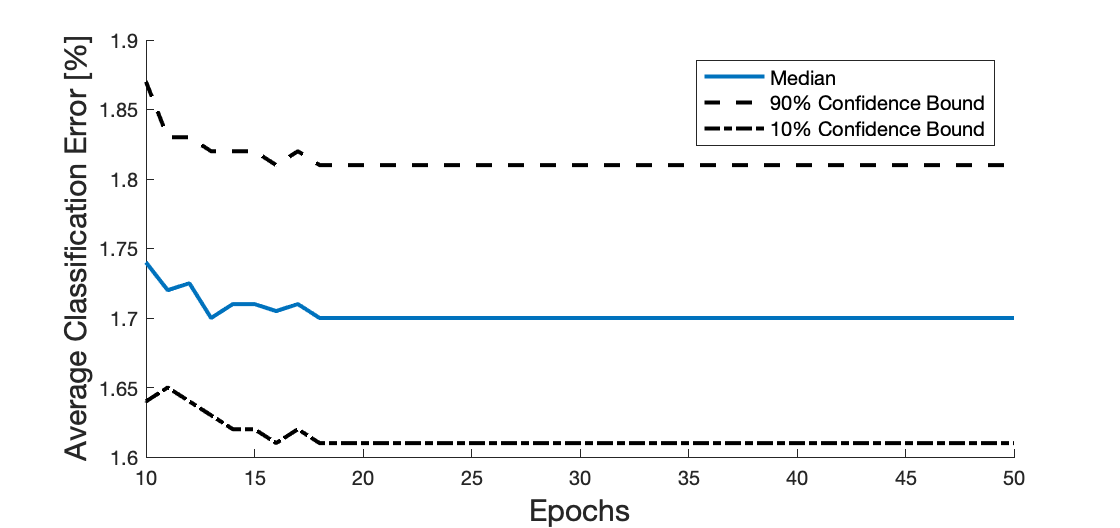}
         %\caption{$100$ Experiments of an attacked network with detection.}
         \caption{Classification Error with Detection (100 Experiments).}
     \end{subfigure}
        %\caption{$100$ Experiments of Label-Flip attack by a single attacker with and without detection, each on a network with $N=5$ agents. The above plots show the $10\%$ and $90\%$ confidence bounds for the average classification error, as well as the average number of successful label-flipping according to the label-flipping function $h(c_i)$ previously described.}
        \caption{Evaluating Label-Flipping Attack Impact: Plots (a) and (b) present the outcomes of 100 experiments without detection, showing classification error and label-flip success rate, respectively. (c) Classification error with detection implemented, offering a comparative perspective. All plots include 10\% and 90\% confidence bounds.}
        \label{fig:ConfidenceBoundsExperimentsLabelFlip}
%\vspace{-\baselineskip}
\end{figure}

\subsection{Example I---Constant-Output Attack}
In this example, the agents are attempting to learn a classifier for the MNIST dataset, while the attacker is aiming to inject a model that consistently outputs the digit $9$. Figure~\ref{fig:ConstantOutputAttackDiffSizes} presents the results of a single experiment of the attack, both with and without the proposed mitigation scheme. We %can 
see in Figure~\ref{fig:ConstantOutputAttackDiffSizes}(a) that the greater the fraction of %malicious
truthful agents, the %faster
more slowly the attack succeeds. However, even with 19 truthful agents, the attack proves successful. But with the proposed scheme, the attacker is detected before it can impact the network, enabling the attacked network to maintain convergence of the model to a good model, as shown in Figure~\ref{fig:ConstantOutputAttackDiffSizes}(b). Figure \ref{fig:ConfidenceBoundsExperiments} displays the accumulated statistics over $100$ experiments, including the $10\%$ and $90\%$ confidence bounds. Figure \ref{fig:ConfidenceBoundsExperiments}(a) shows the statistics of $100$ experiments of an attacked network without detection, and Figure~\ref{fig:ConfidenceBoundsExperiments}(b) illustrates the statistics of $100$ experiments of an attacked network with detection.

\subsection{Example II---Label-Flipping Attack} 
In this example, the agents are attempting to learn a classifier for the MNIST dataset, while the attacker aims to inject a model that flips labels. Consider a label-flipping function $h(c): C\to \widetilde C$, where $C$ is the set of possible labels and $\widetilde C \subseteq C$. For any sample $d_i$ with label $c_i$, the attacker's model will return $h(c_i)$ instead. The label-flipping function used in this example is described in the following table: 
 \begin{center}
 \label{label_flip_table}
\begin{tabular}{ |c| c c c c c c c c c c|}
\hline
 \(c_i\)    & 0 & 1 & 2 & 3 & 4 & 5 & 6 & 7 & 8 & 9 \\
 \hline
 \(h(c_i)\) & 3 & 4 & 7 & 5 & 8 & 0 & 9 & 6 & 2 & 1 \\
 \hline
\end{tabular}
\end{center}

Figure~\ref{fig:ConfidenceBoundsExperimentsLabelFlip} presents the acquired statistics over $100$ experiments, complete with the $10\%$ and $90\%$ confidence bounds. Figure~\ref{fig:ConfidenceBoundsExperimentsLabelFlip}(a) displays the average classification error from $100$ experiments of an attacked network without detection. Figure~\ref{fig:ConfidenceBoundsExperimentsLabelFlip}(b) illustrates the average number of successful label-flippings according to the %previously described 
label-flipping function \(h(c_i)\), and Figure~\ref{fig:ConfidenceBoundsExperimentsLabelFlip}(c) depicts the average classification error from $100$ experiments of an attacked network with detection.

\section{Conclusions}
In this paper, we have presented a robust federated learning algorithm that can operate in the presence of data injection attacks. We have provided conditions for the identification of malicious agents. We have also demonstrated the performance of the proposed technique on various attacks. Detailed proofs of the lemmas as well as bounds on the attacker detection probability and the false-alarm probability are to be presented in an extended version of this work.%, which is to be uploaded to the arXiv.

\balance
\bibliographystyle{IEEEbib}
\bibliography{main}
\end{document}

%% file: macros.tex
\newtheorem{question}{Question}[section]
\newtheorem{coro}{Corollary}[section]

\newcommand{\beq}{\begin{equation}}
\newcommand{\eeq}{\end{equation}}
\newcommand{\bea}{\begin{array}}
\newcommand{\ena}{\end{array}}
\newcommand{\bds}{\begin {itemize}}
\newcommand{\eds}{\end {itemize}}
\newcommand{\bdf}{\begin{definition}}
\newcommand{\blm}{\begin{lemma}}
\newcommand{\edf}{\end{definition}}
\newcommand{\elm}{\end{lemma}}
\newcommand{\bthm}{\begin{theorem}}
\newcommand{\ethm}{\end{theorem}}
\newcommand{\bprp}{\begin{prop}}
\newcommand{\eprp}{\end{prop}}
\newcommand{\bcl}{\begin{claim}}
\newcommand{\ecl}{\end{claim}}
\newcommand{\bcr}{\begin{coro}}
\newcommand{\ecr}{\end{coro}}
\newcommand{\bquest}{\begin{question}}
\newcommand{\equest}{\end{question}}

\newtheorem{lemma}{Lemma}